\documentclass[10pt,twocolumn,letterpaper]{article}

\usepackage{cvpr}
\usepackage{times}
\usepackage{epsfig}
\usepackage{graphicx}
\usepackage{amsmath}
\usepackage{amssymb}
\usepackage{epstopdf}
\usepackage{subfigure}
\usepackage{color}
\DeclareMathAlphabet{\mathcal}{OMS}{cmsy}{m}{n} % fix 

\def \M {\mathcal{M}}

\usepackage[breaklinks=true,bookmarks=false]{hyperref}

\cvprfinalcopy % *** Uncomment this line for the final submission

\def\cvprPaperID{14} % *** Enter the CVPR Paper ID here

\ifcvprfinal\fi % *** Uncomment this line for the final submission
\begin{document}

%%%%%%%%% TITLE
\title{A Counter-Forensic Method for CNN-Based Camera Model Identification}

\author{David G\"uera\\
Purdue University\\
West Lafayette, Indiana\\
%{\tt\small dgueraco@purdue.edu}
% For a paper whose authors are all at the same institution,
% omit the following lines up until the closing ``}''.
% Additional authors and addresses can be added with ``\and'',
% just like the second author.
% To save space, use either the email address or home page, not both
\and
Yu Wang\\
Purdue University\\
West Lafayette, Indiana\\
%{\tt\small wang1317@purdue.edu}
\and
Luca Bondi\\
Politecnico di Milano\\
Milan, Italy\\
%{\tt\small luca.bondi@polimi.it}
\and
%Sri Kalyan Yarlagadda\\
%Purdue University\\
%West Lafayette, Indiana\\
%{\tt\small yarlagad@purdue.edu}
\and
Paolo Bestagini\\
Politecnico di Milano\\
Milan, Italy\\
%{\tt\small paolo.bestagini@polimi.it}
\and
Stefano Tubaro\\
Politecnico di Milano\\
Milan, Italy\\
%{\tt\small stefano.tubaro@polimi.it}
\and
Edward J. Delp\\
Purdue University\\
West Lafayette, Indiana\\
%{\tt\small ace@ecn.purdue.edu}
}

\maketitle
\thispagestyle{empty} % *** Uncomment this line for the final submission

%%%%%%%%% ABSTRACT
\begin{abstract}
%Economies of scale in production and subsequent reductions in the cost of image capturing devices have resulted in an explosion in the number of digital pictures that are generated in a daily basis. 
An increasing number of digital images are being shared and accessed through websites, media, and social applications. Many of these images have been modified and are not authentic. 
%Detecting the camera model used to capture these images can be crucial for criminal investigations and trials. 
Recent advances in the use of deep convolutional neural networks (CNNs) have facilitated the task of analyzing the veracity and authenticity of largely distributed image datasets.  
We examine in this paper the problem of identifying the camera model or type that was used to take an image and that can be spoofed.
Due to the linear nature of CNNs and the high-dimensionality of images, neural networks are vulnerable to attacks with adversarial examples. 
These examples are imperceptibly different from correctly classified images but are misclassified with high confidence by CNNs. 
In this paper, we describe a counter-forensic method capable of subtly altering images to change their estimated camera model when they are analyzed by any CNN-based camera model detector. 
Our method can use both the Fast Gradient Sign Method (FGSM) or the Jacobian-based Saliency Map Attack (JSMA) to craft these adversarial images and does not require direct access to the CNN. 
Our results show that even advanced deep learning architectures trained to analyze images and obtain camera model information are still vulnerable to our proposed method.  
\end{abstract}

%%%%%%%%% BODY TEXT
%%%%%%%%% INTRODUCTION
\section{Introduction}
%The rapid proliferation of inexpensive image capturing devices has driven the widespread diffusion of digital pictures on the web. sharing any type of images through websites and social media is everywhere.
The recent increase in the number of digital images that are being uploaded and shared online has given rise to unique privacy and forensic challenges \cite{Liang2015}. 
Among those challenges, verifying the integrity and authenticity of these widely circulated pictures is one of the most critical and complex tasks \cite{Farid2009, Rocha2011}.

In the last few years, the digital media forensic community has explored several techniques to evaluate the truthfulness of digital images and media \cite{Piva2013, Stamm2013}. 
Due to its multiple applicable scenarios,  research efforts have focused on camera model identification \cite{Kharrazi2004, Filler2008, Kirchner2015,CKM2009,KMM2006}. 
Determining the camera model used to take a picture can be very important in criminal investigations such as copyright infringement cases or  where it is required to identify the authors of pedo-pornographic material. 

Camera model identification can also be considered an important preliminary step to reduce the set of camera instances when we try to detect a unique camera instance rather than just the make and model \cite{Kirchner2015}. 
In addition, being able to identify the camera model by inspecting small image regions is a viable method to uncover manipulation operations that could have been done to the image (\eg splicing) \cite{Swaminathan2008}.

Current camera model identification detectors make use of the fact that each camera model completes a distinctive set of tasks on each image when the device acquires the image. 
Examples of these tasks include the use of different JPEG compression schemes, application of proprietary methods for CFA demosaicing, and ``defects" in the optical image path. 
Due to these characteristic operations, a singular ``footprint" is embedded in each picture. 
This information can be utilized to identify the camera model, and perhaps the exact camera,  that has been used to capture an image or record a video sequence.

Due to the inherent and growing complexity of the image acquisition pipeline of modern image capturing devices, it is a difficult challenge to adequately model the set of operations that a camera has to execute to capture an image. 
Successful attempts that use hand-crafted features to model the traces left by some of these operations can be found in \cite{Filler2008, Choi2006, Bayram2005, Cao2009, Milani2014b,KMM2006,KMD2009,KMD2009b}. 

The use of deep learning techniques for image and video classification tasks \cite{Karpathy2014, LeCun2015, Mas2017} has shown that it is also possible to learn characteristic features that model a problem space directly from the data itself. 
This offers a viable path to leverage the growing amount of available image data. 
These modern approaches are data-driven in that they learn directly from the data rather than imposing a predetermined analytical model.

The data-driven model has recently proved valuable for forensics applications \cite{Buccoli2014, Chen2015, Bayar2016, Xu2016}. 
Initial exploratory solutions targeting camera model identification \cite{Tuama2016b, Bondi2016, Bondi2017} show that it is possible to use CNNs to learn discriminant features directly from the observed known images, rather than having to use hand-crafted features. 
The use of CNNs also makes it possible to capture characteristic traces left by non-linear and hard to model operations present in the acquisition pipeline.

With the introduction of CNNs as detectors for camera model identification, a new vector for counter-forensic attacks is presented for a malevolent skilled individual. 
The idea of counter-forensics was first introduced in \cite{Kirchner2007}, where the authors presented the concept of fighting against image forensics with a practical application, namely a method for resampling an image without introducing pixel correlations. 
An up-to-date survey of the last counter-forensics advances can be found in \cite{Bohme2013}.

Before exploring the vulnerabilities of CNN-based camera model detectors, it is important to note that detectors that rely on hand-crafted features are not immune to similar counter-forensics attacks. 
As explained in \cite{Goljan2010}, digital camera fingerprints are vulnerable to forging. 
In particular, if an attacker obtains access to images from a given camera, they can estimate its fingerprint and ``paste" it into an arbitrary image to make it look as if the image came from the camera with the stolen fingerprint. 
An early attempt to investigate such counter-forensic methods appeared in \cite{Gloe2007}. 

%Let us refocus on the vulnerabilities of the CNNs to adversarial examples attacks. 
As presented in \cite{Szegedy2014}, several machine learning models, including state-of-the-art convolutional neural networks, are vulnerable to adversarial attacks. 
This means that these machine learning models misclassify images that are only slightly different from correctly classified images. 
In many cases, an ample collection of models with different architectures trained on different subsets of the training data misclassify the same adversarial example \cite{Goodfellow2015}.

Although there are techniques such as adversarial training \cite{Szegedy2014} or defensive distillation \cite{Papernot2016b} that can slightly reduce the incidence of adversarial examples in CNN-based detectors, defending against adversarial examples is still an on-going challenge in the deep learning community. 
Adversarial attacks are hard to defend against because they require machine learning models that produce correct outputs for every possible input. 
The imposition of linear behavior when presented with inputs similar to the training data, though desirable, is precisely the main weakness of CNNs \cite{Goodfellow2015}. %Think about this
Due to the massive amount of possible inputs that a CNN can be presented with, it is remarkably simple to find adversarial examples that look unmodified to us but are misclassified by the network. 
Designing a truly adaptive defense against adversarial images remains an open problem.

In this paper, we propose a counter-forensic method to subtly change an image to induce an error in its estimated camera model when analyzed by a CNN-based camera model detector. 
We leverage the recent developments to rapidly generate adversarial images. 
We test our counter-forensic method, using two well established adversarial image crafting techniques \cite{Goodfellow2015, Papernot2016}, against an advanced deep learning architecture \cite{Huang2016} carefully trained on a reference camera model dataset. 
% Our method works at patch level and, when applied to the complete set of patches that compose a given image, allows us to modify the estimated camera model for the individual patches and, consequently, the image. 
Our results show that even modern and properly trained CNNs are susceptible to simple adversarial attacks.
Note that our method only requires access to the predictions of the CNN-based camera model identification detector and does not need access to the weights of the CNN.

%-------------------------------------------------------------------------
%%%%%%%%% CNN-BASED CAMERA MODEL IDENTIFICATION
\section{CNN-Based Camera Model Identification} \label{sec:cnn}

In this section, we provide a brief overview of convolutional neural networks sufficient to understand the rest of this paper and show how they can be used as camera model detectors. For a more detailed description, please refer to one of the several available tutorials in the literature \cite{Kuo2016, Goodfellow2016}.

Convolutional neural networks are a special type of neural networks, biologically inspired by the human visual cortex system, that consist of a very high number of interconnected nodes, or neurons. 
The architecture of a CNN is designed to take advantage of the 2D structure of an input image. 
This is achieved with local connections and tied weights followed by some forms of pooling which results in translation invariant features. 
The nodes of the network are organized in multiple stacked layers, each performing a simple operation on the input.
 
The set of operations in a CNN typically comprises convolution, intensity normalization, non-linear activation and thresholding, and local pooling. 
By minimizing a cost function at the output of the last layer, the weights of the network are tuned so that they are able to capture patterns in the input data and extract distinctive features.

In a CNN, the features are learned using backpropagation \cite{Bengio2009} coupled with an optimization method such as gradient descent \cite{Bottou2010} and the use of large annotated training datasets. 
The shallower layers of the networks usually learn low-level visual features such as edges, simple shapes and color contrast, whereas deeper layers combine these features to identify complex visual patterns.
Finally, fully-connected layers coupled with a softmax layer are commonly used to generate an output class label that minimizes the cost function.
 
For example, in the context of image classification, the last layer is composed of $N$ nodes, where $N$ is the number of classes, that define a probability distribution over the $N$ visual category. 
The value of a given node $p_i, \; i = 1, \dots, N$ belonging to the last layer represents the probability of the input image to belong to the visual class $c_i$. 

To train a CNN model for a specific image classification task we need to define the hyperparameters of the CNN, which range from the sequence of operations to be performed, to the number of layers or the number and shape of the filters in convolutional layers. 
We must also define a proper cost function to be minimized during the training process. 
Finally, a dataset of training and test images, annotated with labels according to the specific task (\eg camera models in our work) needs to be prepared.

\begin{figure}[h]
	\centering
    \includegraphics[width=\linewidth]{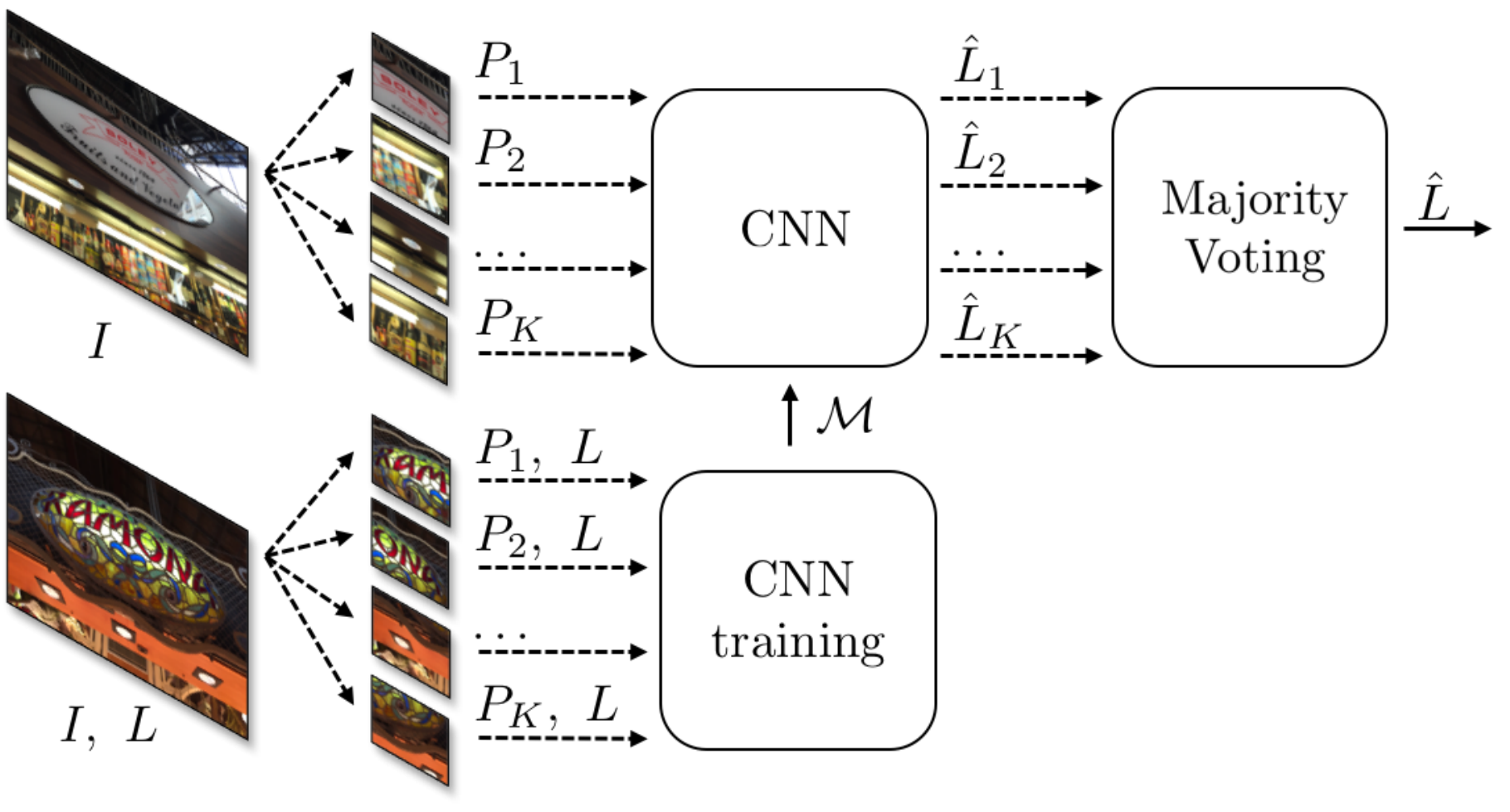}
	\caption{Example of a pipeline for camera model identification. 
	The patches extracted from each training image $I$ (bottom) inherit the same label $L$ of the image. These patches are used in the CNN training process. 
	For each patch $P_k$ from the image $I$ under analysis (top), a candidate label $\hat{L}_k$ is produced by a trained CNN model $\M$. 
	The predicted label $\hat{L}$ for analyzed image $I$ is obtained by majority voting.}
	\label{fig:meth1a}
\end{figure}

Figure~\ref{fig:meth1a} shows an example of a CNN-based pipeline for camera model identification similar to the one presented in \cite{Bondi2016}. 
To train the CNN architecture, we use a given set of training and validation labeled image patches coming from $N$ known camera models. 
For each color image $I$, associated to a specific camera model $L$,  $K$ non-overlapping patches $P_k$, $k \in [1,K]$, of size $32 \times 32$ pixels are randomly extracted. 
Each patch $P_k$ inherits the same label $L$ of the source image. 
As trained CNN model $\M$, we select the one that provides the smallest loss on validation patches.

When a new image $I$ is under analysis, the camera model used to acquire it is estimated as follows. 
A set of $K$ patches is obtained from image $I$ as described above. 
Each patch $P_k$ is processed by CNN model $\M$ in order to assign a label $\hat{L}_k$ to each patch. 
The predicted model $\hat{L}$ for image $I$ is obtained through majority voting on $\hat{L}_k$, $k \in [1,K]$.

%%%%%%%%% PROPOSED METHOD
\section{Proposed Method} \label{sec:meth}

\begin{figure}[h]
	\centering
    \includegraphics[width=\linewidth]{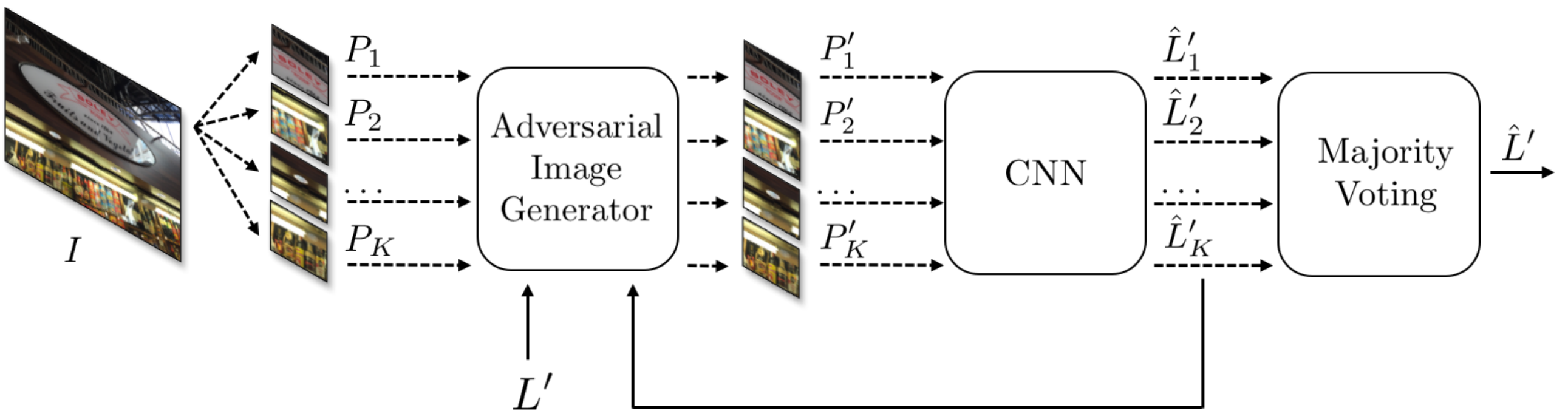}
	\caption{Block diagram of our proposed method.}
	\label{fig:meth1b}
\end{figure}

Figure~\ref{fig:meth1b} shows the block diagram of our proposed counter-forensic method. 
Our method consists of an adversarial image generator module that can be added to a CNN-based camera model evaluation pipeline. In Figure~\ref{fig:meth1b}, we assume a similar structure to the previously presented pipeline in Section \ref{sec:cnn}. 
Our adversarial image generator module takes as input the set of $K$ patches that have been extracted from the image $I$ that is being analyzed. 
When presented with new image patches, our module can work in two different modes. 

In the first operation mode, the adversarial image generator module does an untargeted image manipulation, that is, it does not try to perturb the image patches to produce a specific misclassification class. 
Instead, we use the derivative of the loss function of the CNN with respect to the input image patches to add a perturbation to the images.
The derivative is computed using backpropagation with the labels $\hat{L}'_k$, $k \in [1,K]$ that are given by the CNN detector when it first processes the unmodified image patches.
This procedure is known as the fast gradient sign method (FGSM) \cite{Goodfellow2015}.

In the second operation mode, the adversarial image generator module does a targeted image manipulation. 
In this case, we try to perturb the image patches to produce a specific misclassification class $L'$, different from the true real label $L$ that is associated with the analyzed image $I$ and its associated $P_k$ patches. 
In this mode of operation, we exploit the forward derivative of a CNN to find an adversarial perturbation that will force the network to misclassify the image patch into the target class by computing the adversarial saliency map. 
Starting with an unmodified image patch, we perturb each feature by a constant offset $\epsilon$. 
This process is repeated iteratively until the target misclassification is achieved. 
This procedure is known as the Jacobian-based saliency map attack (JSMA) \cite{Papernot2016}.

We present a detailed overview of both FGSM and JSMA techniques as follows.

%%%%%%%%% FGSM
\subsection{Fast Gradient Sign Method}
In \cite{Goodfellow2015}, the fast gradient sign method was introduced for generating adversarial examples using the derivative of the loss function of the CNN with respect to the input feature vector. 
Given an input feature vector (\eg an image), FGSM perturbs each feature in the direction of the gradient by magnitude $\epsilon$, where $\epsilon$ is a parameter that determines the perturbation size. 
For a network with loss $J(\Theta, x, y)$, where $\Theta$ represents the CNN predictions for an input $x$ and $y$ is the correct label of $x$, the adversarial example is generated as
\begin{equation*}
x^\ast = x + \epsilon \text{sign}(\nabla_xJ(\Theta,x,y))
\end{equation*}
With small $\epsilon$, it is possible to generate adversarial images that are consistently misclassified by CNNs trained using the MNIST and CIFAR-10 image classification datasets with a high success rate \cite{Goodfellow2015}.

%%%%%%%%% JSMA
\subsection{Jacobian-Based Saliency Map Attack}
In \cite{Papernot2016}, an iterative method for targeted misclassification was proposed. 
By exploiting the forward derivative of a CNN, it is possible to find an adversarial perturbation that will force the network to misclassify into a specific target class. 
For an input $x$ and a convolutional neural network $C$, the output for class $j$ is denoted $C_j(x)$. 
To achieve an output of target class $t$, $C_t(x)$ must be increased while the probabilities $C_j(x)$ of all other classes $j \neq t$ decrease, until $t = \arg\max_j$ $C_j(x)$. 
This is accomplished by exploiting the adversarial saliency map, which is defined as
\begin{equation*}
S(x,t)[i]= 
\begin{cases}
    0, \text{if $\frac{\partial C_t(x)}{\partial x_i} < 0$ or $\sum\nolimits_{j \neq t} \frac{\partial C_j(x)}{\partial x_i} > 0$}\\   
    (\frac{\partial C_t(x)}{\partial x_i})|\sum\nolimits_{j \neq t} \frac{\partial C_j(x)}{\partial x_i}|, \text{otherwise}
\end{cases}
\end{equation*}
for an input feature $i$.
Because we work with images in this paper, in our case each input feature $i$ corresponds to a pixel $i$ in the image input $x$.
Starting with a normal sample $x$, we locate the pair of pixels $\{i,j\}$ that maximize $S(x, t)[i] + S(x, t)[j]$, and perturb each pixel by a constant offset $\epsilon$. 
This process is repeated iteratively until the target misclassification is achieved. 
This method can effectively produce MNIST dataset examples that are correctly classified by human subjects but misclassified into a specific target class by a CNN with a high confidence.

%%%%%%%%% IMPLEMENTATION
\subsection{Implementation Details}

To implement our counter-forensic method, we have used the software library \textit{cleverhans} \cite{Papernot2016c}. The library provides standardized reference implementations of adversarial image generation techniques and adversarial training. 
The library can be used to develop more robust CNN architectures and to provide standardized benchmarks of CNNs performance in an adversarial setting. 
As noted in \cite{Papernot2016c}, benchmarks constructed without a standardized implementation of adversarial image generation techniques are not comparable to each other, because a good result may indicate a robust CNN or it may merely indicate a weak implementation of the adversarial image generation procedure.

%-------------------------------------------------------------------------
%%%%%%%%% RESULTS
\section{Experimental Results}

In this section, we evaluate our proposed method and compare the results of the two techniques for generating the adversarial images. 
First, we create a reference dataset specially designed to exploit the traces left by the operations of the acquisition pipeline of different image capturing devices. 
Then, we train an advanced deep learning architecture to have a baseline to compare the accuracy results in the presence of adversarial images. 
Finally, we generate several adversarial image examples to demonstrate the performance of our proposed method. 

%%%% DATASET
\subsection{Experimental Setup}  
As part of DARPA's MediFor Program, PAR Government Systems collected an initial dataset of 1611 images acquired by 10 different camera models, ranging from DSLRs to phone cameras, with a mixture of indoor and outdoor flat-field scenes. We focus on a flat-field image dataset because flat-field images are more difficult to modify without inserting visual distortions due to the absence of texture content.

Throughout the rest of the paper, we refer to this dataset as PRNU-PAR. 
Using the PRNU-PAR dataset, we create a patch dataset, composed by image patches of 32 $\times$ 32 pixels randomly extracted from the original images.
Specifically, 500 patches are uniformly sampled from each original image in the PRNU-PAR dataset, which results in a patch dataset that contains 805,500 patches in total. 
The training, validation and test sets are created following a 70/20/10 split, while we ensure that the patches in each dataset split only contain patches from different images. 

Table \ref{tab:data} shows the statistics of the patch dataset. 
As can be seen, due to the difference in the number of images per camera model class in the PRNU-PAR dataset, our dataset of image patches has an unequal number of patches for each of the camera models.

\begin{table}[h]
\begin{center}
\begin{tabular}{|l|c|c|c|}
\hline
\textbf{Camera Model}       & \textbf{Training}  & \textbf{Validation} & \textbf{Test}  \\ \hline
AS-One       & 90000  & 25500      & 12500 \\ \hline
ES-D5100     & 37500  & 10500      & 5000  \\ \hline
MK-Powershot & 35000  & 10000      & 5000  \\ \hline
MK-s860      & 35500  & 10000      & 5000  \\ \hline
PAR-1233     & 71000  & 20000      & 10000 \\ \hline
PAR-1476     & 107000 & 30500      & 15000 \\ \hline
PAR-1477     & 70000  & 20000      & 9500  \\ \hline
PAR-A015     & 40500  & 11500      & 5500  \\ \hline
PAR-A075     & 26000  & 7000       & 3500  \\ \hline
PAR-A106     & 54000  & 15500      & 7500  \\ \hline
\end{tabular}
\end{center}
\caption{Number of image patches per camera class for each of the different dataset splits.}
\label{tab:data}
\end{table}

\begin{figure}[h]
	\centering
    \includegraphics[width=0.7\linewidth]{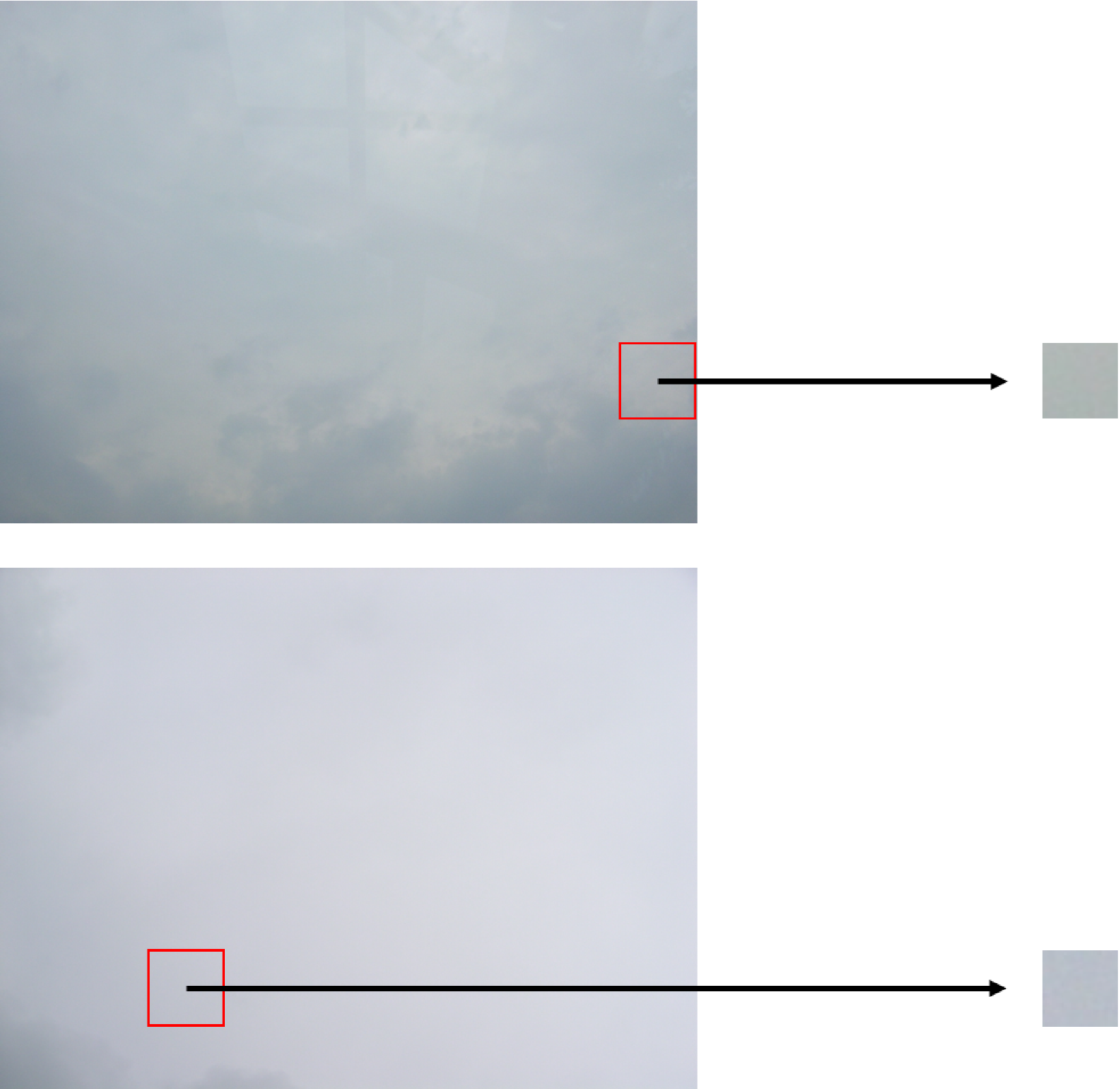}
	\caption{Example of images from the training set of the patch dataset.
	(Top) Image from camera model PAR-A075 and one of the randomly selected patches associated with it. 
	(Bottom) Image from camera model PAR-A106 and one of the randomly selected patches associated with it.}
	\label{fig:patches}
\end{figure}

Figure~\ref{fig:patches} shows a representative example of the images that are present in the PRNU-PAR dataset next to one of their randomly extracted patches. 
In this case, both camera models PAR-A075 and PAR-A106 have been used to capture images of a cloudy sky. 
Other camera models such AS-One or ES-D5100 have taken images of a white screen.
All the image scenes that are captured in the PRNU-PAR dataset are mostly flat and bright. 

As it has been shown in the literature \cite{Filler2008}, these largely uniform images are ideal candidates to be used for the extraction of the ``fingerprint" (\eg the characteristic PRNU noise of the camera model) left in the image by the camera.

%%%% CNN
\subsection{CNN Architecture}

In order to do a fair evaluation of our counter-forensic method, we use a CNN-based camera model detector that has been trained to achieve state-of-the-art accuracy results in the patch dataset.

%We train a DenseNet model \cite{Huang2016}.
CNN architecture designs have tended to explore deeper models.
Networks which can be hundreds of layers deep are now commonplace in the literature. 
This design trend has been motivated by the fact that for many applications such as image classification tasks, an increase in the depth of the CNN architecture translates into higher accuracy performance if sufficient amounts of training data are available.

A first approach to design a CNN architecture may be to simply stack convolutional or fully-connected layers together. 
This naive strategy works initially, but gains in accuracy performance quickly diminish the deeper this kind of architecture becomes. 
This phenomenon is due to the way in which conventional CNNs are trained through backpropogation. 
During the training phase of a CNN, gradient information must be propagated backwards through the network.
This gradient information slightly diminishes as it passes through each layer of the neural network. 
For a CNN with a reduced number of layers, this is not a problem. 
For an architecture with a large number of layers, the gradient signal essentially becomes noise by the time it reaches the first layer of the network again.

The problem is to design a CNN in which the gradient information can be easily distributed to all the layers without degradation.  
ResNets and DenseNets are modern CNN architectures that try to address this problem.

A Residual Network \cite{He2015}, or ResNet is a deep CNN which tackles the problem of the vanishing gradient using a straightforward approach. 
It adds a direct connection at each layer of the CNN.
In previous CNN models, the gradient always has to go through the activations of the layers, which modify the gradient information due to the nonlinear activation functions that are commonly used. 
With this direct connection, the gradient could theoretically skip over all the intermediate layers and be propagated through the network without being disturbed.

A Dense Network \cite{Huang2016}, or DenseNet generalizes the idea of a direct connection between layers. 
Instead of only adding a connection from the previous layer to the next, it connects every layer to every other layer.
For each layer, the feature maps of all preceding layers are treated as separate inputs whereas its own feature maps are passed on as inputs to all subsequent layers.
The increased number of connections ensures that there is always a direct route for the information backwards through the network.
The connectivity pattern of DenseNets yields state-of-the-art accuracies on the CIFAR10 image classification dataset, which is composed by images of 32 $\times$ 32 pixels in size.

Motivated by the accuracy performance of DenseNet in the CIFAR10 dataset and the fact that we also work with image patches of 32 $\times$ 32 pixels, we select a DenseNet model with 40 layers as our CNN camera model detector. 
To prevent the network from growing too wide and to improve the parameter efficiency, we limit the growth rate of the network, this is, the maximum number of input feature-maps that each layer can produce, to $k = 12$. 
To train the CNN, we use the Adam optimizer with a learning rate of $0.0001$ and a batch size of 512 images. After 5 training epochs, we reach a plateau in the accuracy in our validation set. 
Table~\ref{tab:ccn-acc} shows the single patch accuracy results for our training, validation and test splits of the patch dataset. 

\begin{table}[h]
\begin{center}
\begin{tabular}{|l|c|c|c|}
\hline
\textbf{Dataset Split}       & \textbf{Train}  & \textbf{Validation} & \textbf{Test}  \\ \hline 
Accuracy (\%)      & $99.8$  & $98.7$      & $97.7$ \\ \hline
\end{tabular}
\end{center}
\caption{Single patch accuracy results for our training, validation and test splits of the patch dataset.}
\label{tab:ccn-acc}
\end{table}

\subsection{Adversarial Image Generation}

\begin{figure*}[h]
	\centering
    \includegraphics[width=\linewidth]{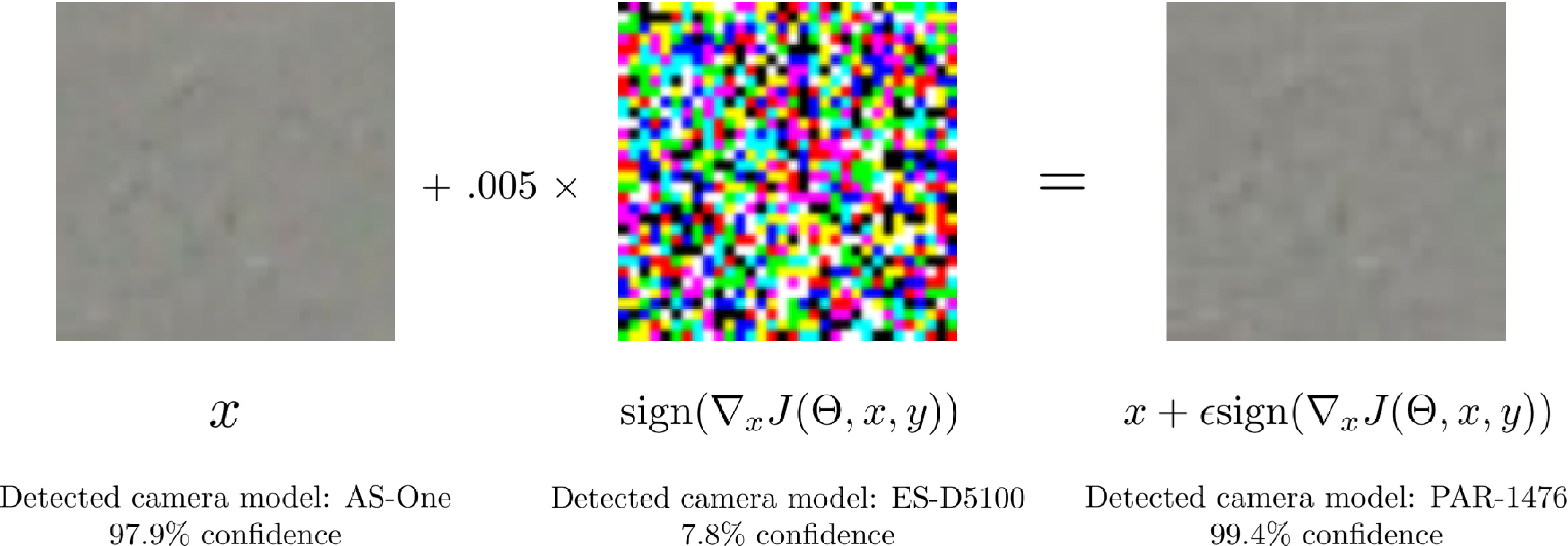}
	\caption{An example of untargeted fast adversarial image generation using FGSM applied to our trained DenseNet model on the patch dataset. 
	By adding an imperceptibly small vector whose elements are equal to the sign of the elements of the gradient of the cost function with respect to the input, we can change DenseNet's classification of the image patch.} 
	\label{fig:fgsm}
\end{figure*}

In order to evaluate the performance of our counter-forensic method, we test the DenseNet model trained on the patch dataset using untargeted attacks with FGSM and targeted attacks with JSMA. 
To properly evaluate our method, we only perturb images from the test split which were correctly classified by our CNN in their original states.

To be clear, what we refer as the average confidence score in this paper is the average value of the probability that is associated with the candidate camera model label for each of the image patches in the test split. 
The probability for each candidate camera model label corresponds with the highest probability value assigned by the softmax layer of our trained DenseNet model.

For untargeted attacks with FGSM, we report in Table \ref{tab:fgsm} the error rate and the average confidence score on the test split of the patch dataset for different values of $\epsilon$ which have been shown to generate high misclassified adversarial images while not producing appreciable visual changes. We find that using $\epsilon = 0.005$ offers the best compromise between error rate and visual changes in the image, causing the trained DenseNet model detector to have a error rate of 93.1\% with an average confidence of 95.3\% on the patch test split. It should be noted that as we increase the value of $\epsilon$, the manipulations become more visually apparent.

Figure \ref{fig:fgsm} shows an example of the adversarial images that our proposed method can generate when we use FGSM. 
The modifications done to the images by FGSM are performed on 32-bit floating point values, which are used for the input of the DenseNet model. 
The gradient computed for Figure \ref{fig:fgsm} uses 8-bit signed integers. 
To publish the sign of the gradient image in the paper, we have done a custom conversion from 8-bit signed integers to 8-bit unsigned integers.
To increase the range of each color channel, we represent the $-1$s values as 0 and the $1$s as 255. For the possible 0's, we have treated them as positive values (they are represented by 255).

\begin{table}[h]
\begin{center}
\begin{tabular}{|c|c|c|}
\hline
\textbf{\shortstack[c]{$\epsilon$ \\ value}} & \textbf{\shortstack[c]{Error \\ rate (\%)}}  &  \textbf{\shortstack[c]{Confidence \\ Score (\%)}}  \\ \hline
0.001        & 91.4  & 97.7       \\ \hline
0.002        & 91.7  & 97.2       \\ \hline
0.003        & 92.2  & 96.7       \\ \hline
0.004        & 92.7  & 95.8       \\ \hline
0.005        & 93.1  & 95.3       \\ \hline
0.006        & 94.1  & 95.1       \\ \hline
0.007        & 94.5  & 94.2       \\ \hline
0.008        & 95.3  & 93.6       \\ \hline
0.009        & 95.9  & 93.0       \\ \hline
0.01         & 96.2  & 92.3       \\ \hline
\end{tabular}
\end{center}
\caption{Error rate and confidence score values of our trained DenseNet model after an untargeted attack with FGSM to the test split with different values of $\epsilon$.}
\label{tab:fgsm}
\end{table}

\begin{table}[h]
\begin{center}
\begin{tabular}{|c|c|c|}
\hline
\textbf{\shortstack[c]{Target \\ Camera Model}} & \textbf{\shortstack[c]{Error \\ rate (\%)}}  &  \textbf{\shortstack[c]{Confidence \\ Score (\%)}}  \\ \hline
AS-One       & 99.5  & 87.7       \\ \hline
ES-D5100     & 99.3  & 88.6       \\ \hline
MK-Powershot & 99.3  & 88.4       \\ \hline
MK-s860      & 99.7  & 88.5       \\ \hline
PAR-1233     & 99.7  & 87.9       \\ \hline
PAR-1476     & 99.4  & 88.1       \\ \hline
PAR-1477     & 99.5  & 88.2       \\ \hline
PAR-A015     & 99.6  & 88.4       \\ \hline
PAR-A075     & 99.3  & 87.8       \\ \hline
PAR-A106     & 99.2  & 87.9       \\ \hline
\end{tabular}
\end{center}
\caption{Error rates and confidence scores of our trained DenseNet model for each possible target camera model after applying a targeted attack with JSMA to the test split.}
\label{tab:jsma}
\end{table}

\begin{figure}[h]
	\centering
    \includegraphics[width=\linewidth]{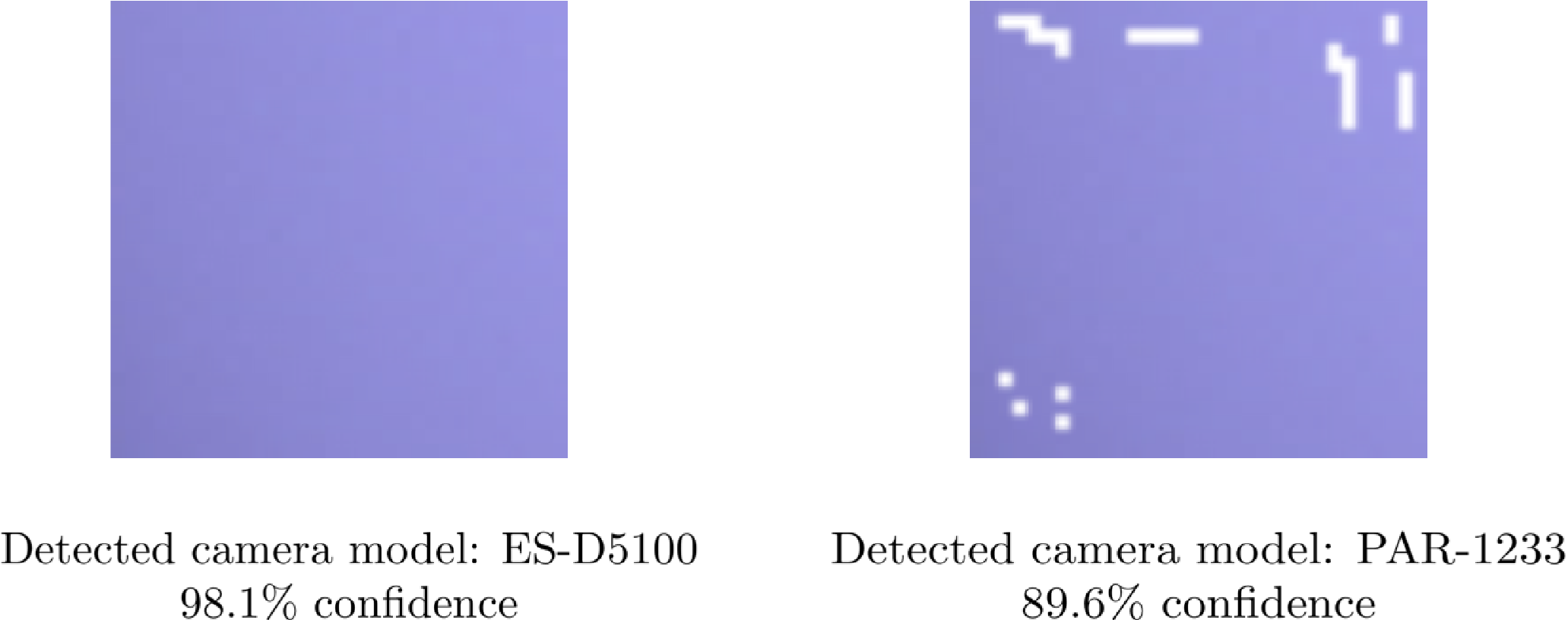}
	\caption{An example of targeted adversarial image generation using JSMA applied to our trained DenseNet model on the patch dataset. 
	(Left) Original image patch correctly classified as ES-D5100.
	(Right) Altered image patch with target camera model PAR-1233}
	\label{fig:jsma}
\end{figure}

For targeted attacks with JSMA, we report in Table \ref{tab:jsma} the error rate and the average confidence score for each possible camera model target class. 
Figure \ref{fig:jsma} shows an example of the images that JSMA allows us to generate when we perform a targeted attack. 
In this case, an image patch captured by camera ES-D5100 that is correctly classified when is analyzed by our trained DenseNet model is manipulated to be misclassified as an image patch that had been generated by camera model PAR-1233.
It is important to appreciate that although JSMA allows us to generate image patches that get misclassified into a specific camera model with high error rates and confidence scores, the modifications that it applies to the images can usually be spotted through visual inspection.
This effect is due to the fact that JSMA crafts the adversarial images by flipping pixels to their minimum or maximum values. 
Because our patch dataset is composed of image patches with mostly flat scene content, the effect can be clearly observed, for example, in the upper corners of the manipulated image patch in Figure \ref{fig:jsma}.

%-------------------------------------------------------------------------
%%%%%%%%% CONCLUSIONS
\section{Conclusions}
This paper described a counter-forensic method to subtly alter images to change their estimated camera model when they are analyzed by a CNN-based camera model detector.
We tested our method on a reference dataset with images from multiple cameras that show highly similar indoor and outdoor scenes. 
The results demonstrate that we can generate imperceptibly altered adversarial images that are misclassified with high confidence by the CNN. 
In the future, we will extend our method to apply it to video sequences and we will explore viable adversarial example detection methods and defense techniques to increase the robustness of CNN-based camera model detectors.

%-------------------------------------------------------------------------
%%%%%%%%% ACKNOWLEDGMENTS
\section{Acknowledgments}
This material is based on research sponsored by the Defense Advanced Research Projects Agency (DARPA) and the Air Force Research Laboratory (AFRL) under agreement number FA8750-16-2-0173. 
The U.S. Government is authorized to reproduce and distribute reprints for Governmental purposes notwithstanding any copyright notation thereon. 
The views and conclusions contained herein are those of the authors and should not be interpreted as necessarily representing the official policies or endorsements, either expressed or implied, of DARPA, AFRL or the U.S. Government.

{\small
\bibliographystyle{IEEEtran}
\bibliography{biblio}
}

\end{document}